\documentclass{article}
\usepackage{ijcai25}

\usepackage{times}
\usepackage{soul}
\usepackage{url}
\usepackage[hidelinks]{hyperref}
\usepackage[utf8]{inputenc}
\usepackage[small]{caption}
\usepackage{graphicx}
\usepackage{amsmath}
\usepackage{amsthm}
\usepackage{booktabs}
\usepackage{algorithm}
\usepackage{algorithmic}
\usepackage[switch]{lineno}
\usepackage[edges]{forest}
\usepackage{multirow}
\usepackage[utf8]{inputenc}

\definecolor{hidden-draw}{RGB}{0,51,102}

\title{A Survey on Ordinal Regression: Applications, Advances, and Prospects}

\author{
Jinhong Wang$^{1,2,3}$
\and
Jintai Chen$^5$\and
Jian Liu$^6$\and
Dongqi Tang$^6$\and
Danny Z. Chen$^7$\And
Jian Wu$^{2,3,4}$
\affiliations
$^1$College of Computer Science \& Technology Zhejiang University, Hangzhou 310012, China \\
$^2$State Key Laboratory of Transvascular Implantation Devices of The Second Affiliated Hospital, Zhejiang University School of Medicine, Hangzhou 310009, China \\ 
$^3$Zhejiang Key Laboratory of Medical Imaging Artificial Intelligence \\ 
$^4$School of Public Health Zhejiang University, Hangzhou 310058, China \\
$^5$HKUST (Guangzhou) \quad
$^6$AntGroup \quad
$^7$University of Notre Dame \\
\emails
\{wangjinhong, wujian2000\}@zju.edu.cn,
\{jtchen721, dongqi.tdq\}@gmail.com, \\
rex.lj@antgroup.com,
dchen@nd.edu
}

\begin{document}

\maketitle

\begin{abstract}
    Ordinal regression refers to classifying object instances into ordinal categories. Ordinal regression is crucial for applications in various areas like facial age estimation, image aesthetics assessment, and even cancer staging, due to its capability to utilize ordered information effectively. More importantly, it also enhances model interpretation by considering category order, aiding the understanding of data trends and causal relationships. Despite significant recent progress, challenges remain, and further investigation of ordinal regression techniques and applications is essential to guide future research. In this survey, we present a comprehensive examination of advances and applications of ordinal regression. By introducing a systematic taxonomy, we meticulously classify the pertinent techniques and applications into three well-defined categories based on different strategies and objectives: \textit{Continuous Space Discretization}, \textit{Distribution Ordering Learning}, and \textit{Ambiguous Instance Delving}. This categorization enables a structured exploration of diverse insights in ordinal regression problems, providing a framework for a more comprehensive understanding and evaluation of this field and its related applications. To our best knowledge, this is the first systematic survey of ordinal regression, which lays a foundation for future research in this fundamental and generic domain.
\end{abstract}

\section{Introduction}
Ordinal regression, also known as ordinal classification, aims to solve classification problems in which categories are not independent but instead follow a natural order. Many real-world phenomena naturally provide ordinal data, such as facial age estimation (see Figure 1). In the past, researchers ignored the common order nature of ordinal regression tasks and built models independently~\cite{ranjan2015unconstrained,qawaqneh2017deep}. 

In recent years, researchers connected these tasks as ordinal regression and strived to develop techniques for this general problem formulation, whose importance and 
significance are embodied in three aspects: {\bf (1) Broader cross-domain applications.} Ordinal regression can be broadly applied in many fields, like image quality measure~\cite{wang2023deep}, monocular depth estimation~\cite{wang2024scalable}, and even some extremely critical areas such as cancer staging~\cite{albuquerque2021ordinal,toledo2022grading}. Ordinal regression provides a unified perspective for the study of these tasks, so as to learn common and effective underlying logic and solution. {\bf (2) Ordered information utilization.} Unlike normal classification, ordinal regression takes into account the natural ordering of categories in the response variable. This allows for more nuanced and accurate modeling when the order of categories carries meaningful information~\cite{liu2019probabilistic}, as well as training a universal model to learn ordered information from different ordinal regression tasks jointly. {\bf (3) Improved model interpretation.} Since ordinal regression considers order relationships between categories, the output results of the model are more explanatory, which can help researchers better understand the causal relationships and trends behind the data~\cite{du2024teach}. In recent years, the progress of ordinal regression research has been remarkable. Various methods have emerged and provided many valuable insights. However, there are still many issues that need to be resolved. It is highly desirable to delve into the technical nuances, inner insights, and diverse application scenarios of ordinal regression, examining their common discoveries and distinctive contributions to provide guidance for future research, which is the purpose of this survey.

 \begin{figure*}[t]
\centering
\includegraphics[width=0.99\textwidth]{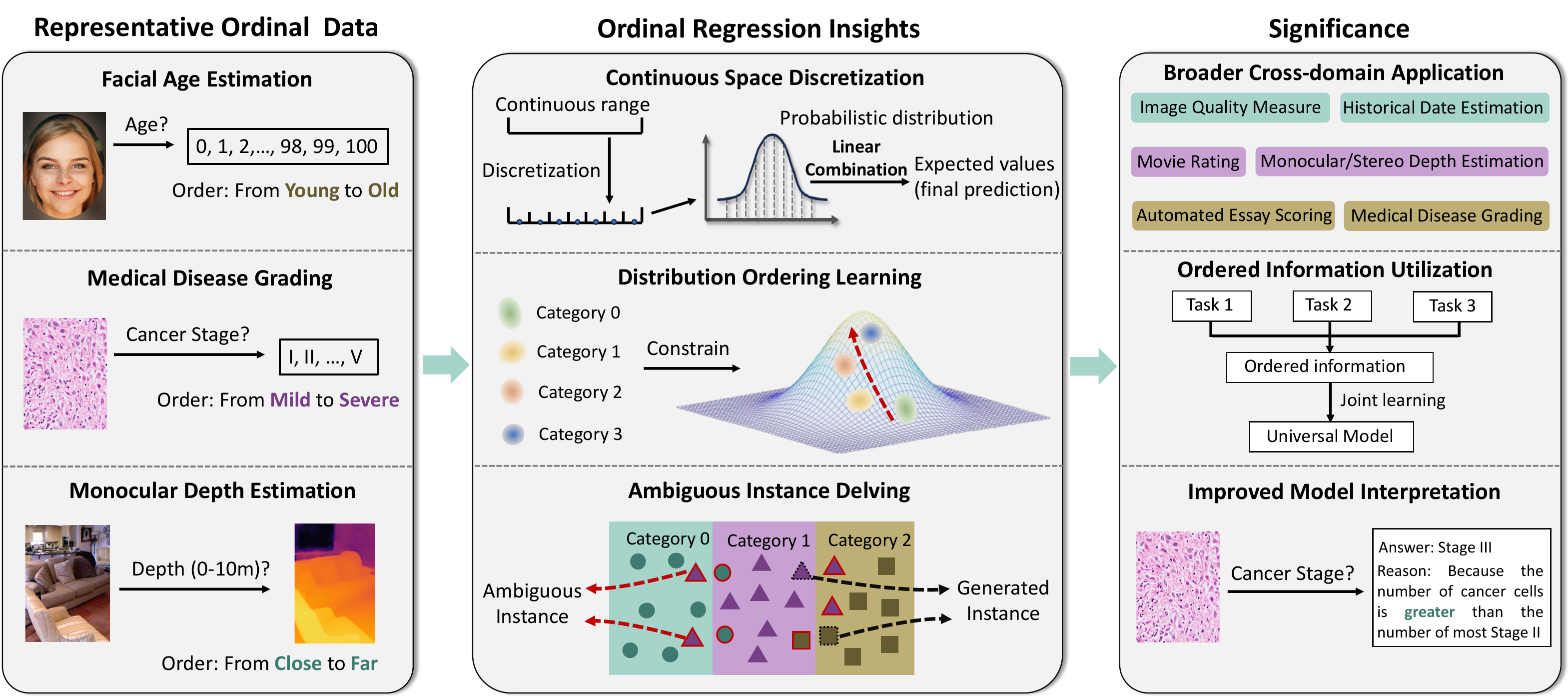}
\caption{Demonstrating the applications, insights, and significances of ordinal regression. Many real-world phenomena naturally provide ordinal data such as the age of a person, the level of a disease, and the depth of the radar. For the prediction of ordered data, ordinal regression was explored recently mainly in three aspects: \textit{Continuous Space Discretization}, \textit{Distribution Ordering Learning}, and \textit{Ambiguous Instance Delving}. Importantly, the study of ordinal regression has three main significances.  (1) Ordinal regression methods can be applied to broader applications since the underlying ordered logic is similar. (2) Ordinal regression methods can fully utilize meaningful order information in ordinal data and even there is a potential to train a universal model on all ordered data. (3) Ordinal regression methods can be more interpretable since they consider order relationships between categories which may contain causal information and trends behind the data.
}
\label{fig1}
\end{figure*}

To systematically understand and analyze the development of ordinal regression, this paper classifies the research and applications 
following a taxonomy that categorizes the \textit{modeling rationale} into three paradigms: \textit{Continuous Space Discretization}, \textit{Distribution Ordering Learning}, and \textit{Ambiguous Instance Delving}.  This classification allows for a comprehensive overview of the current state-of-the-art AI in ordinal regression tailored to common tasks, highlighting each category’s unique insights. 
By categorizing these insights based on their applications and algorithms, we aim to provide an extensive examination that facilitates easier comprehension and identification of existing research gaps.


\textit{Continuous Space Discretization} focuses on transforming ordinal regression into a deep classification task via continuous space discretization. This transformational insight enlightens the field of ordinal regression more than its performance effectiveness. Many recent advances still explored this general idea to solve the ordinal regression problem. We delve into the subcategories of image-level estimation and pixel-level estimation based on spatial hierarchy differences, reflecting the diversity of ordinal regression applications and methods.
 \textit{Distribution Ordering Learning} addresses the main specificity of ordinal regression that differs from simple classification tasks: Labels are in order. This category discusses how deep learning (DL) methods facilitate ordinal relationship modeling between categories via distribution ordering learning. It underscores the development from traditional labeling and probability distribution modeling to a more modern, CLIP-based approach. \textit{Ambiguous Instance Delving} explores another key issue that needs to be addressed in practical applications of ordinal regression: Models often mis-predict samples to adjacent categories due to blurred boundaries and confusing patterns. This category covers the discussion of different solutions for better distinguishing ambiguous instances across adjacent categories, including rank-based, reference comparison, and generative models.

To our best knowledge, this is the first systematic survey of ordinal regression, which lays a foundation for future research and applications in this fundamental and generic domain. The scope of this survey extends beyond mere categorization
of current techniques and applications. It also explores the potential future trajectories of ordinal regression research, contemplating how ongoing advancements might unfold. The discussion encompasses not only technological aspects but also the challenges of deploying advanced ordinal regression methods in more and wider applications.

\section{Taxonomy}
The taxonomy aims to group ordinal regression methods in various applications with similar insights into the same category, facilitating in-depth investigation in future studies. We classify techniques and applications into three
distinct categories based on insights and problems of ordinal regression: \textit{Continuous Space Discretization}, \textit{Distribution Ordering Learning}, and \textit{Ambiguous Instance Delving}. A visual presentation of this taxonomy is shown in Figure 2.

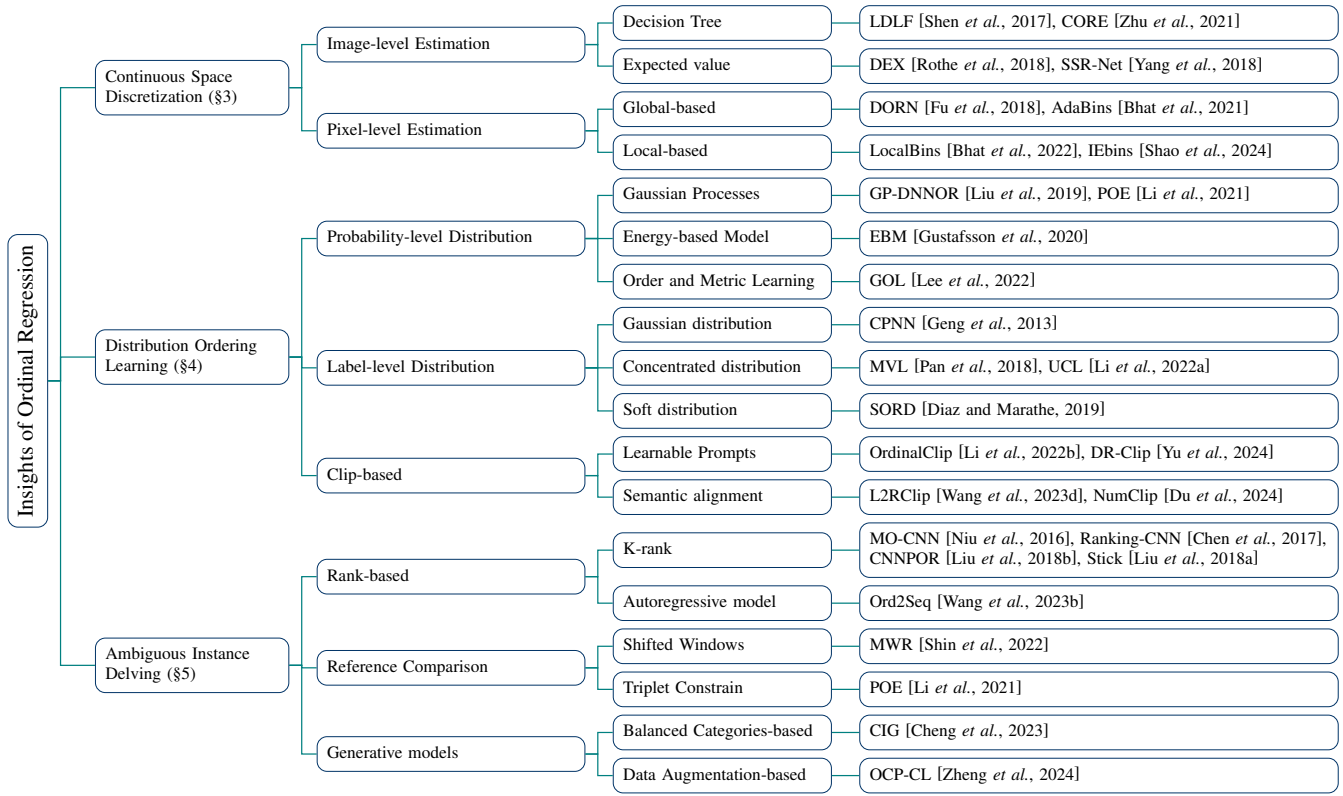
\begin{figure*}[t]
    \centering
    \resizebox{\textwidth}{!}{
        \begin{forest}
            forked edges,
            for tree={
                grow=east,
                reversed=true,
                anchor=base west,
                parent anchor=east,
                child anchor=west,
                base=left,
                font=\small,
                rectangle,
                draw=hidden-draw,
                rounded corners,
                align=left,
                minimum width=4em,
                edge+={teal, line width=0.5pt},
                s sep=3pt,
                inner xsep=4pt,
                inner ysep=3pt,
                ver/.style={rotate=90, child anchor=north, parent anchor=south, anchor=center},
            },
            where level=1{text width=7em,font=\scriptsize,}{},
            where level=2{text width=10em,font=\scriptsize,}{},
            where level=3{text width=8em,font=\scriptsize,}{},
            where level=4{text width=18.5em,font=\scriptsize,}{},
            [
                Insights of Ordinal Regression, ver
                [
                    Continuous Space \\ Discretization (\S\ref{csp})
                    [
                        Image-level Estimation 
                        [   
                            Decision Tree
                                [
                                    LDLF~\cite{shen2017label}\text{,} 
                                    CORE~\cite{zhu2021convolutional}
                                ]
                        ]
                        [  
                            Expected value
                                [
                                    DEX~\cite{rothe2018deep}\text{,} 
                                    SSR-Net~\cite{yang2018ssr}
                                ]
                        ]
                    ]
                    [
                        Pixel-level Estimation
                        [
                            Global-based 
                            [
                                DORN~\cite{fu2018deep}\text{,} AdaBins~\cite{bhat2021adabins}
                            ]
                        ]
                        [
                            Local-based 
                            [
                                LocalBins~\cite{bhat2022localbins}\text{,} 
                                IEbins~\cite{shao2024iebins}
                            ]
                        ]
                    ]
                ]
                [
                    Distribution Ordering \\ Learning (\S\ref{dol})
                    [
                        Probability-level Distribution
                        [
                            Gaussian Processes
                            [
                                GP-DNNOR~\cite{liu2019probabilistic}\text{,} 
                                POE~\cite{li2021learning}
                            ]
                        ]
                        [
                            Energy-based Model
                            [
                                EBM~\cite{gustafsson2020energy}
                            ]
                        ]
                        [   
                            Order and Metric Learning
                            [
                                GOL~\cite{lee2022geometric}
                            ]
                        ]
                    ]
                    [
                        Label-level Distribution
                        [   
                            Gaussian distribution
                            [
                                CPNN~\cite{geng2013facial}
                            ]    
                        ]
                        [
                            Concentrated distribution
                            [
                                MVL~\cite{pan2018mean}\text{,} 
                                UCL~\cite{li2022unimodal}
                            ]
                        ]
                        [
                            Soft distribution 
                            [
                                SORD~\cite{diaz2019soft}
                            ]
                        ]
                    ]
                    [
                        Clip-based
                        [
                            Learnable Prompts
                            [
                                OrdinalClip~\cite{li2022ordinalclip}\text{,} 
                                DR-Clip~\cite{yu2024clip}
                            ]
                        ]
                        [
                            Semantic alignment
                            [
                                L2RClip~\cite{wang2023learning}\text{,} 
                                NumClip~\cite{du2024teach}
                            ]
                        ]
                    ]
                ]
                [
                    Ambiguous Instance \\ Delving (\S\ref{aid})
                    [
                        Rank-based
                        [   
                            K-rank
                            [   
                                MO-CNN~\cite{niu2016ordinal}\text{,} 
                                Ranking-CNN~\cite{chen2017using}\text{,} \\
                                CNNPOR~\cite{liu2018constrained}\text{,} 
                                Stick~\cite{liu2018ordinal}\
                            ]
                        ]
                        [
                            Autoregressive model
                            [
                                Ord2Seq~\cite{wang2023ord2seq}
                            ]           
                        ]
                    ]
                    [
                        Reference Comparison
                        [
                            Shifted Windows
                            [
                                MWR~\cite{shin2022moving}
                            ]
                        ]
                        [
                            Triplet Constrain
                            [
                                POE~\cite{li2021learning}
                            ]
                        ]
                    ]
                    [
                        Generative models
                        [
                            Balanced Categories-based
                            [
                                CIG~\cite{cheng2023robust}
                            ]
                        ]   
                        [
                            Data Augmentation-based 
                            [
                                OCP-CL~\cite{zheng2024enhancing}
                            ]
                        ]
                    ]
                ]
            ]
        \end{forest}}
    \caption{\label{fig:taxonomy_intro}
    A taxonomy of ordinal regression with representative examples.
    }
    \vspace{-0.6em}
    \label{fig:tax}
\end{figure*}

\subsection{Continuous Space Discretization}
The earliest ordinal regression algorithm was to pose ordinal regression as a deep classification problem by dividing a continuous range of values into discrete candidate values, called continuous space discretization. The final regression values are calculated by computing expected values over the softmax-normalized probabilities of discrete candidate values. Based on different spatial hierarchies, the methods of continuous space discretization can be divided into the following two subcategories.

\noindent
{\bf Image-level Estimation} usually collects only one label for an entire image. For example, the facial age estimation task can be solved in a classification manner by discretizing the entire range of age values followed by a softmax expected value
refinement. 

\noindent
{\bf Pixel-level Estimation} requires a more dense prediction for the entire image. For example, monocular depth estimation aims to predict the depth of every single pixel in an image. In such a scenario, the requirements for discretization are much more than image-level estimation. Researchers may need to consider whether to model each pixel independently.

\subsection{Distribution Ordering Learning}
Distribution ordering learning highlights the importance of inter-class ranking relationships, which are a key feature that distinguishes ordinal regression from unordered (usual) classification tasks. This topic explores how to inject ordinal relations into model learning, which we summarize into the following three types.

\noindent
{\bf Logit-level Distribution.} Such methods advocate ordering learning based on logit features, assuming that the features of each category should conform to a certain order in the latent space -- that is, a category ranking order. By imposing ordering distribution constraints on latent logit features, the model is encouraged to learn the differences in categories.

\noindent
{\bf Label-level Distribution.} Many studies explored constraining ordinal relationships among categories by seamlessly incorporating metric penalties into ground truth label representations. By performing different label distribution processing, the model can easily apply various metric learning and ordering learning schemes between categories. 

\noindent
{\bf CLIP-based Pre-training.} Recent years have witnessed great success of pre-training general language representations. Especially with the emergence of CLIP, many CLIP-powered methods have been proposed for visual classification tasks as well as ordinal regression. Pertaining to CLIP, models can learn concepts from the language to acquire inter-class values and ordering knowledge.

\subsection{Ambiguous Instance Delving}
Unlike general classification tasks, it is often challenging to distinguish ambiguous instances among adjacent categories due to their confusing data patterns or blurred boundaries in ordinal regression tasks. Various methods have been proposed to deal with the problem of distinguishing ambiguous samples, which we coalesce into three main strategies.


\noindent
{\bf Ranking-based Models}. As a most typical strategy, ranking-based methods use multiple binary classifiers to distinguish adjacent categories one by one. This type of method is simple and basic but may have a high computational cost. 

\noindent
{\bf Instance Referencing.} Instance reference methods perform triplet training to enhance ambiguous instance prediction. For each target sample, the model samples two instances of adjacent categories as references to constrain the target sample modeling. As a trade-off, the model needs to perform two additional feature extraction processes each time.

\noindent
{\bf Generative Models}. These models utilize a useful insight to solve the ambiguous instance delving problem. Adjacent categories are difficult to distinguish, in part because boundaries are difficult to determine. These methods generate as many samples on the boundary as possible to strengthen the boundary determination and distinction of adjacent categories.

\section{Continuous Space Discretization}
\label{csp}
A typical continuous space discretization based solution treats ordinal regression tasks as classification tasks, which is useful in many visual applications, regardless of whether the labels are integers or decimals, and for image level annotation or pixel level annotation. In this section, we cover two key aspects in the applications of continuous space discretization based on different spatial hierarchies: \textit{Image-level Estimation} and \textit{Pixel-level Estimation}.

\subsection{Image-level Estimation}
Common applications of image-level estimation tasks include facial age estimation, image aesthetic assessment, etc. These tasks have only image-level labels which often are integers and ordered. Traditional methods treat these tasks as simple classification tasks and focus only on network architectures, including CNN-based~\cite{ranjan2015unconstrained,qawaqneh2017deep} and Attention-based models~\cite{zhang2019fine}. In contrast, as the earliest application scenario of ordinal regression, multiple continuous space discretization methodologies came into being, with \textit{Decision Tree} and \textit{Expected Value} as two main approaches.

\noindent
{\bf Decision Tree.} Decision Tree~\cite{robinson1965machine}, as a typical classification method, can handle complex classification tasks recursively based on a certain logic. Taking the order properties of ordinal regression tasks as a logical basis, LDLF~\cite{shen2017label} presents label distribution learning forests based on differentiable decision trees for facial age estimation; it defines a distribution-based loss function for the forests to be learned jointly in an end-to-end manner. CORE~\cite{zhu2021convolutional} further integrates differentiable decision trees with a convolutional neural network (CNN) where CNN is used to jointly learn the binary classifiers of the decision trees to obtain a more global feature representation.

\noindent
{\bf Expected Value.} If an ordinal regression task is treated as a classification task and the softmax maximum is taken as the final prediction, then this is no different from a simple classification task. Dex~\cite{rothe2015dex,rothe2018deep} first proposed to compute a softmax expected value of each age category as the final prediction for facial age estimation. This is based on continuity and distribution of the age range. In this case, age estimation can be viewed as a piece-wise regression, or alternatively, as a discrete classification with multiple discrete value labels. Dex proved that the larger the number of categories is, the smaller the discretization error gets for the regressed signal. Inspired by the Dex approach of performing regression by calculating the expected values, SSR-Net~\cite{yang2018ssr} further takes a coarse-to-fine strategy and performs multi-class classification with multiple stages for facial age estimation. Each stage is responsible only for refining the decision of its previous stage for more accurate age estimation. Furthermore, SSR-Net assigns a dynamic range to each age class by allowing it to be shifted and scaled according to the input face image.

\subsection{Pixel-level Estimation}
Pixel-level estimation tasks mainly include depth estimation, in which the labels of one image cover the depth value of each pixel. In most scenarios, depth values are usually in a non-integer continuous space, and thus a traditional approach was mainly based on a pure regression algorithm. In recent years, it has been found that depth estimation can be treated as ordinal regression and transformed into a classification task to achieve high accuracy. Depending on the overall or individual processing of pixels, these methods can be grouped into two types: \textit{Global-based} and  \textit{Local-based}.

\begin{table*}[t]
\centering
\scalebox{0.95}{
\begin{tabular}{l|cccc}
\toprule
Applications                               & Dataset Name                             & Dataset Size & Label Range & Category Groups Number \\ \midrule
                                           & Adience~\cite{wang2023ord2seq}              &      26,580        &     [0, 60+]        &    8 groups             \\
                                           & MORPH II~\cite{shin2022moving} &      55,000        &     [16, 77]        &  Int  \\
                                           & FG-Net~\cite{shin2022moving} &      1,002        &   [0, 69]          &     Int  \\
                                           & CLAP2015~\cite{shin2022moving}                 &  4,691     &  [3, 85]  & Int \\
                                           & UTK~\cite{shin2022moving}                          & 20,000                &     [0, 116]        & int \\
                                           & CACD~\cite{shin2022moving}                          &      160K        &     [14, 62]       & Int \\        
\multirow{-7}{*}{Facial Age Estimation}    & IMDB-WIKI~\cite{shin2022moving}                &     500,000         &  [0, 100]       &    Int\\ \midrule
                                           & Aesthetics~\cite{wang2023ord2seq}      &   13,706           &     [1, 5]        &     5 groups        \\
                                           & AADB~\cite{kong2016photo}                  &     10,000         &     [1, 10]        &   10 groups              \\
                                           & AVA~\cite{kong2016photo}           &    250,000           &    [78, 549]         & Int                \\
\multirow{-4}{*}{Aesthetics Assessment}    & TAD66K~\cite{he2022rethinking}         &      66K        &  [0, 10]           &      -           \\ \midrule
                                           & APTOS~\cite{yu2024clip}      &      13,000        &  [0, 4]           &     5 groups            \\
                                           & DeepDR~\cite{yu2024clip}     &       2,000       &      [0, 3]       &        4 groups         \\
\multirow{-3}{*}{Diabetic Retinal Grading} & EyePACS~\cite{cheng2023robust}      &       35,126       &    [0, 4]         &   5 groups             \\ \midrule
\multirow{-1}{*}{Historical Image Estimation} & HCI~\cite{wang2023ord2seq}      &   265           &  [1930, 1970]           &     5 groups            \\ \midrule
                                            &  KITTI~\cite{shao2024iebins}      &   42,949           &  [0, 80]           &     -            \\ 
                                            &  NYU Depth v2~\cite{shao2024iebins}      &   24,231            &  [0, 10]           &     -            \\ 
\multirow{-3}{*}{Monocular Depth Estimation} &  SUN RGB-D~\cite{shao2024iebins}      &   5,050           &  [0, 10]           &     -            \\ 
\bottomrule
\end{tabular}
}
\caption{Popular datasets and applications of ordinal regression. `Int' means that each integer in the range of labels represents a category; `-' means that the labels are decimals, and thus there is no exact number of categories. }
\label{tab1}
\end{table*}

\noindent
{\bf Global-based.} Earlier methods that applied ordinal regression for depth estimation performed the same strategy for each pixel and thus could be considered global-level processing. Similar to image-level continuous space discretization, by discretizing the depth range into several intervals as candidate categories, the final depth values can be computed as the expected value of each depth candidate. DORN~\cite{fu2018deep} first introduces a spacing-increasing discretization (SID) strategy to discretize depths into several intervals (bins) and recast depth network learning as an ordinal regression problem. The discretization process used in DORN is pre-determined and is the same for each pixel or even each image. Following DORN, AdaBins~\cite{bhat2021adabins} proposes to set an adaptive discretization strategy for each image and to predict the final depth values as a linear combination of bin centers. Although the discretization strategy of every image is independent, each pixel of the same image is still a whole and is processed globally.

\noindent
{\bf Local-based.} For each image, the strategy of pixel discretization with global consistency has a disadvantage: the value distribution of each pixel may be different. LocalBins~\cite{bhat2022localbins} is aware of this and proposes to predict depth distributions of local neighborhoods at every pixel, instead of predicting global depth distributions. Effective improvements facilitated the use of this strategy in subsequent methods~\cite{shao2024iebins}. IEbins~\cite{shao2024iebins} proposes elastic local bins to gradually compress the range of depth value candidates to improve accuracy. DAR~\cite{wang2024scalable} further integrates autoregressive models with ordinal regression for local depth distribution modeling.

\section{Distribution Ordering Learning}
\label{dol}
Distribution ordering learning aims to impose constraints on the features or predicted labels to conform to a certain distribution or order. This is the prior knowledge brought about by the nature of the ordinal regression task itself. This section explores three directions of modeling ordinal relationships between categories: \textit{Logit-level Distribution}, \textit{Label-level Distribution}, and \textit{CLIP-based Pre-training}.

\subsection{Probabilistic-level Distribution}
Many methods choose to constrain ordinal distributions on latent features or probabilistic levels, with the intention to focus more on objective, image-level supervision. Depending on the constraints or methods applied, we summarize these methods into three types: \textit{Gaussian Process}, \textit{Energy-based Models}, and \textit{Order and Metric Learning}.

\noindent
{\bf Gaussian Process.}  GP-DNNOR~\cite{liu2019probabilistic} proposes to adapt traditional Gaussian process regression for ordinal regression tasks by using both conjugate and non-conjugate ordinal likelihoods.  POE~\cite{li2021learning} proposes to learn probabilistic ordinal embeddings which represent each data as a multivariate Gaussian distribution rather than a deterministic point in the latent space. Further, an ordinal distribution constraint is proposed to exploit the ordinal nature of regression and is integrated into the Gaussian distribution process.

\noindent
{\bf Energy-based Models.} EBM~\cite{gustafsson2020energy} proposes a general and conceptually simple ordinal regression method with a clear probabilistic interpretation. Specifically, EBM creates an energy-based model of the conditional target density $p(y|x)$, using a deep neural network to predict the un-normalized density from $(x, y)$. This model of $p(y|x)$ is trained by directly minimizing the associated negative log-likelihood, approximated using Monte Carlo sampling. 

\noindent
{\bf Order and Metric Learning.} 
GOL~\cite{lee2022geometric} proposes geometric order learning for ordinal regression. GOL imposes two geometric constraints, the order constraint and metric constraint, on the latent probabilistic features in a created embedding space. In this latent embedding space, the direction and distance between objects represent order and metric relations between their ranks. The order constraint compels objects to be sorted according to their ranks, while the metric constraint makes the distance between objects reflect their rank difference. 

\subsection{Label-level Distribution}
The work of learning from a label distribution has achieved promising results on ordinal regression tasks such as facial age and head pose estimation. Adjusting the distribution of the ground-truth labels can change the model's training target, which must follow the order constraint. In this section, we 
summarize different label distributions on ordinal regression tasks into three types:  \textit{Gaussian Distribution}, \textit{Concentrated Distribution}, and \textit{Soft Label Distribution}.

\noindent
{\bf Gaussian Distribution.} Facial age estimation is a typical ordinal regression task since the faces at close ages look quite similar, as aging is a slow and smooth process. Inspired by this observation, CPNN~\cite{geng2013facial} regards each face image as an instance associated with a Gaussian label distribution, instead of taking each face image as an instance with one label (age). In this way, one face image can contribute to not only the learning of its chronological age, but also the learning of the ordinal relationships with adjacent ages.

\noindent
{\bf Concentrated Distribution.} A recent work~\cite{li2022unimodal} suggests that learning an adaptive label distribution in ordinal regression tasks should follow two principles.  First, the probability corresponding to the ground-truth should be the highest in label distribution.  Second, the probabilities of neighboring labels should decrease with the increase of distance away from the ground-truth, i.e., the distribution is unimodal. Under these principles, Concentrated Distribution is the most appropriate method. MVL~\cite{pan2018mean} first proposes to learn a concentrated label distribution as well as the mean value close to the ground-truth label. UCL~\cite{li2022unimodal} further proposes adaptive label distribution learning with unimodal-concentrated loss to make the predicted label distribution maximize at the ground-truth and vary according to the predicted uncertainty.

\noindent
{\bf Soft Label Distribution.}  For ordinal regression tasks, it is crucial to classify each class correctly while learning adequate inter-class ordinal relationships.  SORD~\cite{diaz2019soft} presents a novel label distribution strategy that constrains these relationships among categories by seamlessly incorporating metric penalties into ground-truth label representations. Specifically, SORD converts data labels into soft probability distributions that pair well with common categorical loss functions (e.g., cross-entropy) where soft labels can lean towards each ordinal category stronger or weaker according to their continuous distance metric likelihoods.

\subsection{CLIP-based Pre-training}
Ordinal relation supervision based on CLIP pre-training is a new direction emerged in recent years. Unlike other methods that highlight distribution constraints, the CLIP-based approach relies mainly on semantic information of the text to teach the model a sense of order. Thanks to the success of large models, this kind of method has gradually become a main research direction and is worth further exploration.

\noindent
{\bf Learnable Prompts.} CLIP~\cite{radford2021learning} proposes to learn representations from image-text pairs. Experimental results have shown that CLIP is much more efficient at zero-/few-shot transfer, which demonstrate the strong power of language. Encouraged by the excellent performance of CLIP, many CLIP-powered methods have been proposed for vision classification tasks~\cite{patashnik2021styleclip,zhang2021tip}. CoOp~\cite{zhou2022learning} proposes learnable vectors to automatically model a prompt’s context words without spending a significant amount of time on words tuning. Inspired by CoOp, OrdinalCLIP~\cite{li2022ordinalclip} first introduces CLIP into ordinal regression. Similar to CoOp, OrdinalCLIP concatenates word embedding and learnable prompt context to learn ordinal relation properties of language prototypes automatically and extract rank concepts from the rich semantic CLIP latent space.

\noindent
{\bf Semantic Alignment.} Learnable prompt-based CLIP has limitations in preserving semantic alignment in latent feature space.
L2RCLIP~\cite{wang2023learning} proposes to perform ordering learning and semantic alignment simultaneously based on CLIP. L2RCLIP constructs a series of rank-specific sentences as prompts for semantic alignment, and performs a token-level attention mechanism to enhance the ordering relations of prompts.  After concatenating these with global context prompts in the word embedding space, a contrastive loss is used to refine this CLIP feature space. In a recent advance, NumCLIP~\cite{du2024teach} disassembles the exact image to number-specific text matching problem into coarse classification and fine prediction stages. NumCLIP constructs a series of phrases to map age relationships for semantic alignment at the coarse stage, and performs fine-grained cross-modal ranking-based regularization loss to keep both semantic and ordinal alignment in CLIP’s feature space at the fine stage.

\section{Ambiguous Instance Delving}
\label{aid}
Ambiguous instance delving, a main problem that limits the performance of ordinal regression tasks, has largely been ignored by previous research. This problem arises from confusing data patterns and blurred boundaries in ordinal regression tasks, which make it challenging to solve as most prediction errors on ambiguous samples fall into adjacent categories. Recently, it gradually attracted research attention and lots of related works have emerged.

\subsection{Ranking-based Models}
Rank-based models are among the earliest methods for ordinal regression, which distinguish adjacent ambiguous categories based on multiple binary classifiers and determine the final category by combining predictions. 

\noindent
{\bf K-rank.} The K-rank method~\cite{frank2001simple} is one of the most popular approaches for ordinal regression, in which $K - 1$ binary classifiers are trained to transform ordinal regression into a series of binary classification sub-problems that can focus on distinguishing adjacent categories. A study in~\cite{li2006ordinal} combines mathematical analysis based on the K-rank method to better inject ordinal relationships into adjacent ambiguous category distinguishing. MO-CNN~\cite{niu2016ordinal} and Ranking CNN~\cite{chen2017using} further used trained CNNs as K-rank binary classifiers and the binary outputs are aggregated for the final prediction.

\noindent
{\bf Autoregressive Models.} A drawback of the K-rank method is that classifiers may be redundant and independent. Ord2Seq~\cite{wang2023ord2seq} introduces a novel autoregressive model for ordinal regression, in which only $\log_{2}{K}$ classifiers are used. By transforming ordinal regression into an autoregressive sequence prediction task, the prediction objective is decomposed into a series of recursive binary classification steps to better distinguish adjacent categories in a process of progressive elaboration. In this process, each classification is refined based on the results of the previous classification, therefore establishing connections between classifiers.

\subsection{Instance Reference}
This is a kind of sampling-based approach, which can constrain the feature learning of the model and focus on distinguishing adjacent categories simultaneously by sampling two additional instances of adjacent categories as references.

\noindent
{\bf Triplet Constraint.} POE~\cite{liu2019probabilistic} proposes an order constraint method based on triples of adjacent categories.  When sampling a triplet and its probabilistic embedding, POE aims to preserve a distance metric constraint in the latent space to learn the relationships and differences between adjacent categories through the other two instances.

\noindent
{\bf Shifted Windows.} MWR~\cite{shin2022moving} designs a search window to constrain the range of predictions. Moreover, MWR iteratively selects two reference instances to better distinguish adjacent categories and adjust the range of the search window based on the comparison results. As the search window becomes smaller, the prediction range shrinks, making the prediction more accurate.

\subsection{Generative Models}
In practical scenarios, ambiguous instance prediction may become more challenging when the categories are unbalanced or the samples on the boundaries are few. Generative models are promising to mitigate this situation, by generating as many ambiguous samples on the boundaries as possible or data augmentation for data balance. Thanks to the capability to synthesize image
details with higher fidelity, generative adversarial networks (GANs)~\cite{goodfellow2020generative} have been widely adopted to reconstruct high-quality images.

\noindent
{\bf Balanced Categories.} CIG~\cite{cheng2023robust} shows that the class imbalance and category overlap issues can seriously hinder the performance of ordinal regression models. As a result, the performance on minority categories is often unsatisfactory. 
 CIG proposes to generate extra training samples with specific labels near category boundaries, and the sample generation is biased toward the less-represented categories via GAN. In this way, the boundary of adjacent samples is easier to determine, which is conducive to distinguishing adjacent categories, and the improvement is more significant for minority categories.

\noindent
{\bf Data Augmentation.} In a recent advance,  OCP-CL~\cite{zheng2024enhancing} explores the application of contrastive learning to improve ordinal regression models, but finds that strong data enhancement can hinder performance improvement. Thus, OCP-CL proposes to utilize GAN models to generate images with different styles but containing desired ordinal content information as data augmentation. Through generative data augmentation, discriminant information of adjacent categories can be effectively increased to improve the performance of the model.

\section{Datasets, Challenges, and Future Prospects}

\subsection{Datasets}
Table~\ref{tab1} presents a collection of popular datasets for main applications of ordinal regression. We mainly introduce applications in five areas: Facial Age Estimation, Aesthetics Assessment, Diabetic Retinal Grading, Historical Image Estimation, and Monocular Depth Estimation. The first four are for image-level prediction tasks while the last one is for a pixel-level prediction task. We also provide details of each dataset, including the dataset name, dataset size, range of ground-truth labels, and the number of category groups. Notably, `Int' means that each integer in the range of labels represents a category; `-' means that the labels are decimals, and thus there is no exact number of categories. Labels of the datasets have already been grouped by pre-processing. These datasets provide the foundation for ordinal regression research and also reflect the universality of ordinal regression.

\subsection{Challenges}

\noindent
{\bf Ambiguous Samples Distinguishing.}
A main challenge in ordinal regression still lies in ambiguous sample distinguishing problems across adjacent categories. For example, in medical applications, disease grading standard is often determined subjectively by doctors, which leads to ambiguous boundaries~\cite{wang2023transformer}. On the other hand, the characteristics of adjacent samples can be very similar, making ambiguous samples more confusing and difficult to distinguish.

\noindent
{\bf Inconsistent Label Distributions.} The labels of different datasets are often not consistent, even for the same application task. Taking the facial age estimation task as an example, while some datasets have 8 categories, each category covering a different age range, some other datasets even have only 5 categories. Hence, it is challenging to train a general model. Previous methods always train separate models for different datasets, and the use of models is very inefficient.

\noindent
{\bf Out-of-domain Generalization.} Currently, independent models need to be trained for different application tasks. For example, models for age estimation have poor zero generalization performance in aesthetic evaluation and medical grading. Even though all such applications are ordinal regression tasks, the generalizability issue still belongs to the out-of-domain (OOD) problem, and this OOD problem is difficult to mitigate by existing methods.

\subsection{Future Prospects}

\noindent
{\bf VLM-assisted Order Learning.} 
With the emergence of CLIP-based methods, it is promising that vision-language models (VLMs) can be used for order learning in the future. A feasible approach is to introduce the human type of thinking on ordinal logic based on the chain of thought. For example, we can use a VLM with chain of thought to decompose ordinal regression predictions into multiple steps: first predicting a rough range, then further predicting a refined range, and finally predicting the sole category.  At present, there is no work that explores the utilization of existing VLMs for zero-shot generalization validation or chain of thought to decompose the ordinal regression thinking process.

\noindent
{\bf Minority Categories Recognition.} For some ordinal regression applications, the recognition of minority categories is still worth exploring. For example, in medical disease grading, the number of severe samples is often very small compared to the normal and mild grades, making severe samples be predicted into milder categories~\cite{cheng2023robust}. For the problem of predicting such minority samples in adjacent categories, it is promising to mitigate the challenge by long-tail learning or more biased ordering learning.

\noindent
{\bf Ordinal Regression Foundation Models.} In recent years, large language models (LLMs) have dominated various fields and spawned corresponding expert foundation models. It is well worth developing a foundation model for ordinal regression. At present, no foundation model has been proposed for ordinal regression, which is a valuable topic worth exploring in future research. A potential benefit is that such a foundation model can unify all downstream applications of ordinal regression to achieve excellent zero-shot or few-shot generalization performance. A key challenge to developing such a foundation model is how to accommodate the general order relation modeling paradigm that can be applied in all ordinal regression tasks.

\section*{Contribution Statement}
Jian Wu and Danny Z. Chen are the corresponding authors.

\bibliographystyle{named}
\bibliography{ijcai25}

\end{document}